\documentclass[10pt]{article}
\usepackage{wws2023} 
\usepackage{url}
\usepackage{latexsym}
\usepackage{amsmath, amsthm, amsfonts}
\usepackage{algorithm, algorithmic}  
\usepackage{graphicx}
\usepackage{lipsum} 
\usepackage{fancyhdr}
\usepackage[headheight=1in,margin=1in]{geometry}
\usepackage{hyperref}
\usepackage{multicol}
\usepackage{multirow}
\usepackage{subcaption}

\newcommand{\eg}{\textit{e.g.}}

\newcommand{\Eg}{\textit{E}.\textit{g}.}

\let\svthefootnote\thefootnote

\title{WikiWeb2M: A Page-Level Multimodal Wikipedia Dataset}

\author{
  Andrea Burns \\
  Boston University\footnotemark \\\And
 Krishna Srinivasan \\
  Google \\\And
  Joshua Ainslie  \\
  Google \\\And
  Geoff Brown \\
  Google \\\AND
  Kate Saenko \\
  FAIR, Boston University \\\And
  Bryan A. Plummer \\
  Boston University \\\And
  Jianmo Ni \\
  Google \\\And
  Mandy Guo \\
  Google
  }

\begin{document}
\maketitle

\thispagestyle{fancy}
\let\thefootnote\relax\footnotetext{\hspace{-1.75mm}$^{*}$Work was done during an internship at Google.}
\addtocounter{footnote}{-1}\let\thefootnote\svthefootnote

\begin{abstract}
Webpages have been a rich resource for language and vision-language tasks. Yet only pieces of webpages are kept: image-caption pairs, long text articles, or raw HTML, never all in one place. Webpage tasks have resultingly received little attention and structured image-text data underused. To study multimodal webpage understanding, we introduce the Wikipedia Webpage 2M (WikiWeb2M)\footnote{Data is readily available at \url{https://github.com/google-research-datasets/wit/blob/main/wikiweb2m.md}} suite; the first to retain the full set of images, text, and structure data available in a page. WikiWeb2M can be used for tasks like page description generation, section summarization, and contextual image captioning. 
\end{abstract}


\section*{Introduction}
Webpages are multimodal, structured content which can been used for pretraining and fine-tuning. Large scale noisy datasets scraped from the web have been used to pretrain large language or contrastive models~\cite{2020t5,jia2021scaling}. Downstream tasks built from webpages have included instruction following, image captioning, news captioning, image-sentence retrieval, and image-article retrieval~\cite{htmlllm,Biten2019GoodNE,Tan2022NewsStoriesIA}. Yet little prior work has studied tasks to evaluate multimodal webpage understanding itself. 

Many classification and generation problems could be studied with webpages: taxonomic webpage classification, webpage retrieval, web image captioning, and webpage summarization. However, to date there is no open source, multimodal dataset that retains all webpage content.  \Eg, the Wikipedia Image Text (WIT) dataset~\cite{wit} does not keep HTML structure and misses out on many text sections, as shown in Table~\ref{tab:datacomparetrain}. Unified text, image, and structure data would allow for greater study of multimodal content understanding with many-to-many text and image relationships. As a result, we propose the new Wikipedia Webpage (WikiWeb2M) dataset of over 2M pages, which unifies webpage content to include all text, images, and their location (\eg, section index) in one example. Table~\ref{tab:aggregstat} (left) includes the number of pages, sections, and images, along with sample counts for downstream tasks.

\begin{figure*}[!ht]
\centering
    \includegraphics[scale=0.135]{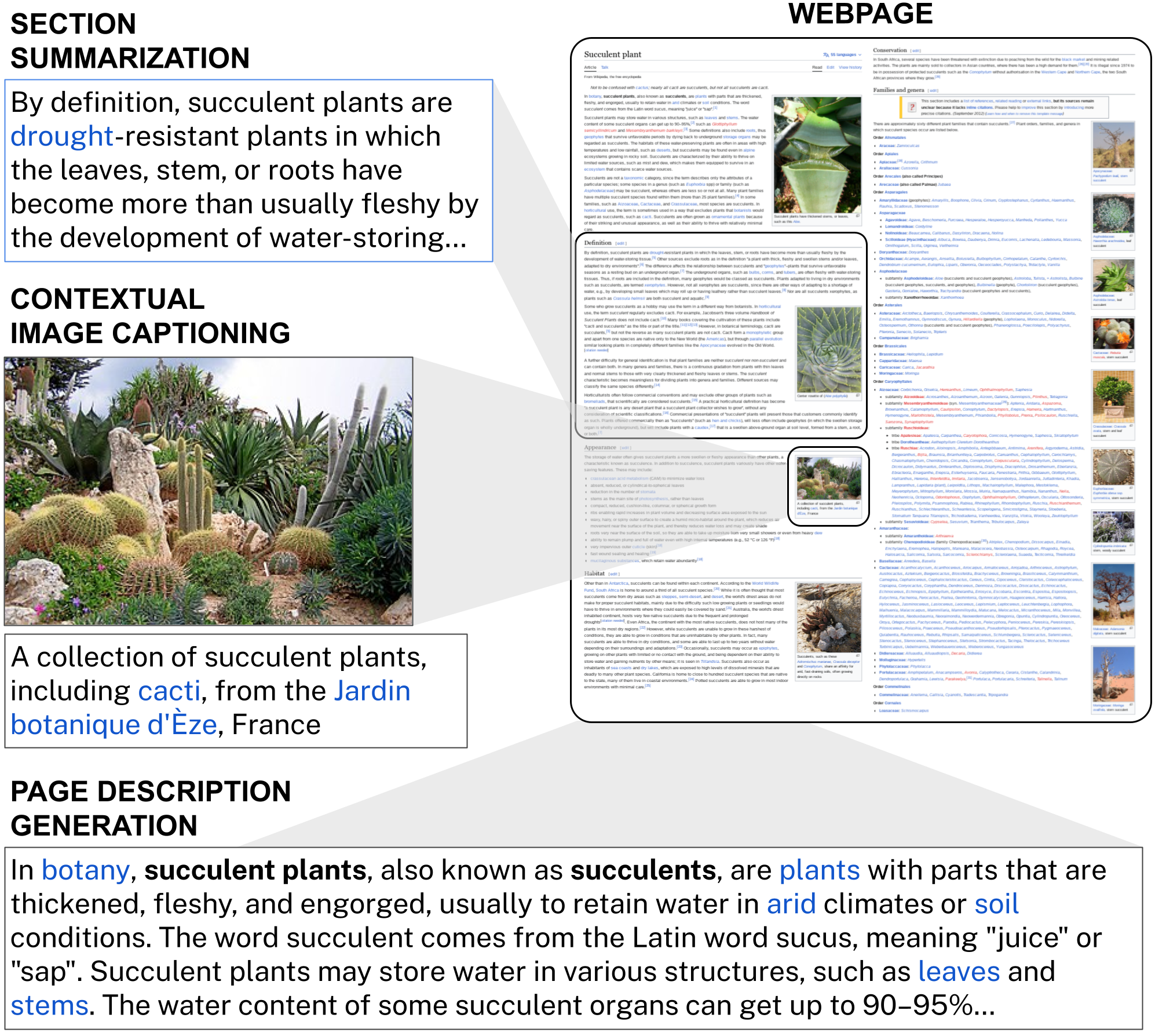}
    \includegraphics[scale=0.135]{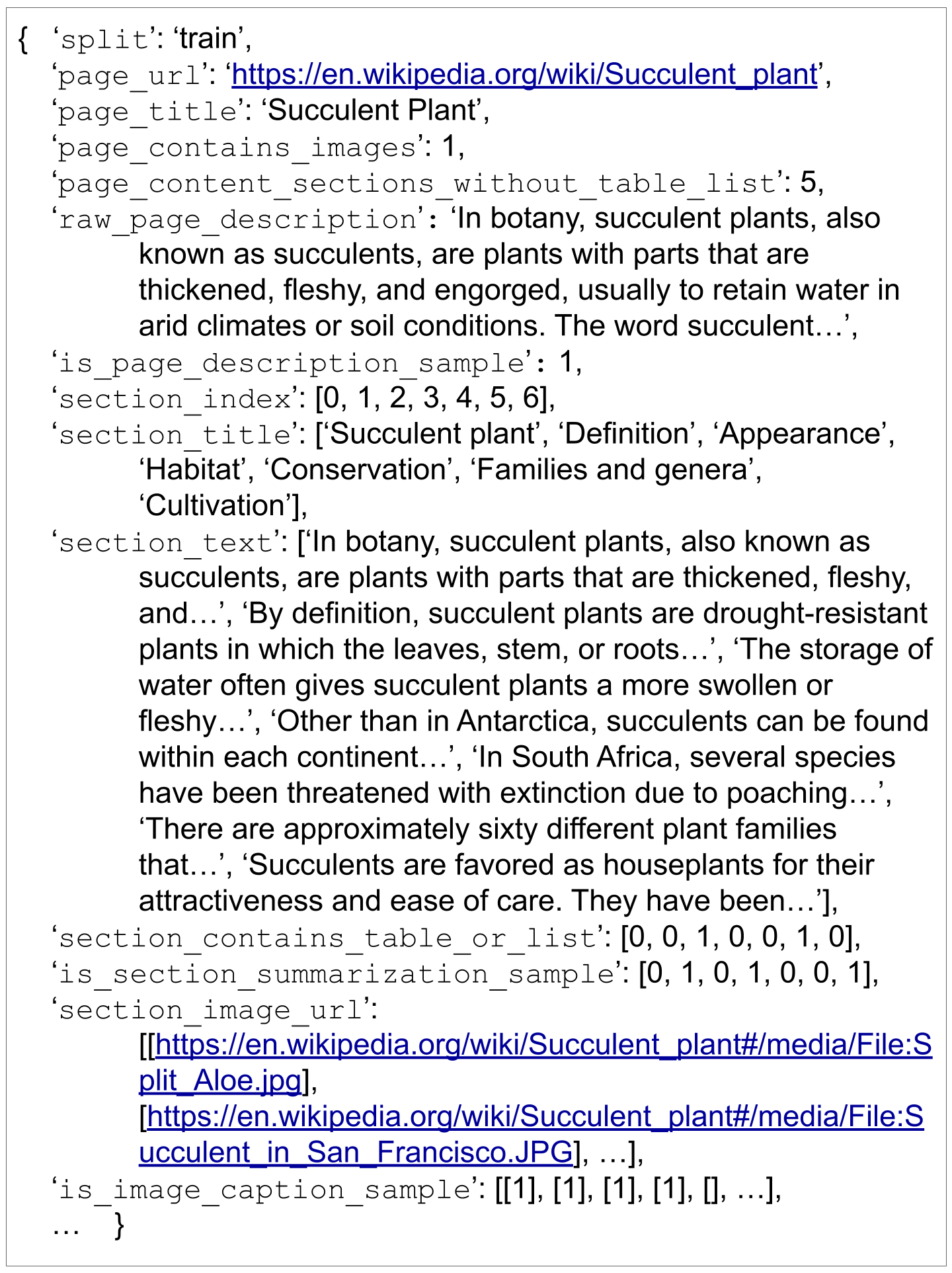}
    \caption{Example tasks and data samples from WikiWeb2M. On the left we show how our dataset provides a unified webpage sample that contains all text, image, and structure, enabling new tasks like page description generation. For image captioning and section summarization, remaining page text and images provide useful context, aiding task performance. On the right we show the WikiWeb2M page sample for the same Wikipedia article on succulents; we only include a subset of fields due to space. \Eg, the WikiWeb2M sample also contains the image alt-text, attribution and reference descriptions, along with other metadata, but it is not illustrated on the right.} 
    \label{fig:motivation}
\end{figure*}

Figure~\ref{fig:motivation} (left) shows how one webpage can be used for page description, section summarization, and contextual captioning. 
These tasks can improve interaction with web content, \eg, a page description may provide a user who is blind more agency by allowing them to preview content before listening to the entire body with a screen reader~\cite{screenreader}. On top of aiding assistive technology, tasks like contextual image captioning and section summarization can be used for modern content generation, as there is growing interest in providing multimodal snippets from the web~\cite{wiki2story}.

\section*{The WikiWeb2M Dataset}
WikiWeb2M is created by rescraping the $\sim$2M English articles in WIT. 
Each webpage sample includes the page URL and title, section titles, text, and indices, images and their captions, and more; see Figure~\ref{fig:motivation} (right). This differs from WIT which defined individual samples as image-caption pairs with additional metadata (\eg, originating section title). 

We shuffle the WIT webpages to define a random 1.8M/100K/100K train/val/test split. Table~\ref{tab:aggregstat} (left) shows the number of pages, sections, and images in our dataset after additional processing. In particular, we only retain content sections (\eg, not the ``See Also'' section). For images, we keep JPEG and PNG and require the dimensions be greater than 1px to allow for a greater diversity of images to be included (\eg, icons)\footnote{We release image URLs, where they can be fetched.}. We include metadata on image dimensions to allow for additional filtering.

In Table~\ref{tab:datacomparetrain}, we report the number of sections and images compared to the English subset of WIT. We add nearly 1M total images to the dataset by keeping the images on a webpage regardless of whether they have image captions.
We break down section counts by type: structural, heading, text, image, and both text and image. Structural and heading sections do not contain immediate section text (the former have subsections).
For heading sections, the section content either linked to a different article, was empty, or only had tables.
A notable 6.8M text sections are in WikiWeb2M, none of which were available in WIT. 

\section*{The WikiWeb2M Tasks}
\label{sec:tasks}
We now describe WikiWeb2M's suite of multimodal generation tasks and task data processing. 
Table~\ref{tab:aggregstat} (left) shows data statistics and (right) downstream task performance when using T5 and ViT base models~\cite{2020t5,dosovitskiy2020vit}.

\smallskip
\noindent\textbf{Page Description Generation}
The goal is to generate a description of a page given the rest of the webpage's image, text, and structure. We use the Wikipedia-provided page descriptions for each article. 
We retain a page if the description has at least five words. A small subset of Wikipedia pages are lists\footnote{For example, \url{https://en.wikipedia.org/wiki/List_of_mammals_of_the_United_States}}; we remove pages that explicitly have ``list\_of'' in their URL or fewer than two rich content sections.

\smallskip
\noindent\textbf{Section Summarization}
The goal is to generate a sentence that highlights the section's content given images and (non-summary) text in the section and other context sections. We take advantage of the leading sentence bias and use the first sentence of a section its pseudo summary. In a small pilot, a majority of human annotators also deemed the first sentence as a reasonable summary.
A section serves as a target section if it has at least five sentences, contains neither a table nor list, and is not the root section. We filter out the root because the root (first) section is often the page description.

\smallskip
\noindent\textbf{Contextual Image Captioning} \cite{contextualcap} proposed Wikipedia image captioning given the image's webpage context. With WikiWeb2M, we can now utilize the entire webpage context for the image instead of just the section it originally came from.
We only allow target images to be those from WIT to ensure quality captions.
Following prior work, we also use the reference description as the ground truth caption to be generated and require it must have at least three words. But, we do not input the attribution description, as it often contains large overlap with the reference description.

\smallskip
\noindent\textbf{Results} Table~\ref{tab:aggregstat} (right) shows results for each task. For contextual image captioning and section summarization we verify that WikiWeb2M's additional sections (compared to only inputting the target section for image captioning or summarization) improve task performance; page description generation is only made possible with our dataset. 
\vspace{-2mm}

\bibliographystyle{wws2023} 
\bibliography{references}

\clearpage
\setlength{\tabcolsep}{5.75pt}
\begin{table*}[ht]
\centering
\begin{tabular}{|l|cccccc|cc|}
\hline
\multirow{2}{*}{\textbf{Dataset}} & \multicolumn{6}{c|}{\textbf{\# Webpage Sections}} &\multicolumn{2}{c|}{\textbf{\# Images}} \\
\cline{2-9}
& Structural & Heading & Text & Image & Both & Total & Unique & Total \\
\hline 
WIT (En) & - & - & - &  199,872 & 2,847,929 & 3,047,801 & 3,660,211 & 4,955,835 \\
WikiWeb2M & 731,394 & 686,376 & 6,817,950 & 221,523 & 3,236,254 & 11,693,497 & 4,438,642 & 5,940,431 \\ 
\hline
\end{tabular}
\caption{Comparison of WikiWeb2M to WIT. We report the aggregate counts over all splits. WikiWeb2M and WIT (English subset) contain the same webpages.}
\label{tab:datacomparetrain}
\end{table*}

\setlength{\tabcolsep}{4.75pt}
\begin{table*}[ht]
\centering
\begin{tabular}{|l|ccc|}
\hline
\textbf{WikiWeb2M Statistic} & \textbf{Train} & \textbf{Val} & \textbf{Test} \\
\hline 
\# Pages & 1,803,225 & 100,475 & 100,833 \\
\# Sections & 10,519,294 & 585,651 & 588,552\\
\# Total Images & 5,340,708 & 299,057 & 300,666 \\
\# Task Samples & & & \\
  \hspace{2mm} Page Description & 1,435,263 & 80,103 & 80,339 \\
  \hspace{2mm} Section Summarization & 3,082,031 & 172,984 & 173,591 \\
  \hspace{2mm} Contextual Captioning & 2,222,814 & 124,703 & 124,188 \\
    \hline
\end{tabular}\hspace{4mm}
\begin{tabular}{|l|ccc|}
\hline
\textbf{Downstream Task} & \textbf{B} & \textbf{R} & \textbf{C} \\
\hline
  Page Description & 14.00 & 38.50 & 81.49 \\
  \hline
  Section Summarization & & & \\
\hspace{2mm} Target Section Only & 8.90 & 27.82 & 60.20 \\
\hspace{2mm} WikiWeb2M & 10.12 & 29.43 & 69.89 \\
\hline
  Contextual Captioning & & & \\
  \hspace{2mm} Target Section Only & 10.92 & 36.21 & 148.53\\
\hspace{2mm} WikiWeb2M & 11.84 & 37.69 & 158.19 \\
    \hline
\end{tabular}

\caption{Statistics and experimental results on the WikiWeb2M dataset. On the left we report the number of pages, sections, and images in the source WikiWeb2M dataset. Below, we report the number of samples for three task datasets that we generate from WikiWeb2M with additional processing: page description generation, section summarization, and contextual image captioning. On the right we report the task performance achieved with T5 and ViT base models (metrics include BLEU-4 (B), ROUGE-L (R), and CIDEr (C)).}
\label{tab:aggregstat}
\end{table*}

\clearpage
\end{document}